\documentclass[11pt]{article}
\usepackage[utf8]{inputenc}
\usepackage[T1]{fontenc}
\usepackage{lmodern}
\usepackage[margin=1in]{geometry}
\usepackage{amsmath,amssymb}
\usepackage{amsthm}
\usepackage{graphicx}
\usepackage{booktabs}
\usepackage{hyperref}
\usepackage{xcolor}
\usepackage{natbib}
\usepackage{enumitem}
\usepackage{placeins}
\usepackage{xurl}
\usepackage{microtype}
\hypersetup{colorlinks=true, linkcolor=blue!70!black, citecolor=blue!70!black, urlcolor=blue!70!black}
\newtheorem{proposition}{Proposition}
\newtheorem{lemma}{Lemma}

\title{Removable and Irreducible:\\ A Token-Cost Ledger for the Multilingual Tokenization Tax}
\author{Madhulatha Mandarapu\thanks{madhulatha@samyama.ai} \and Sandeep Kunkunuru\thanks{sandeep@samyama.ai}}
\date{VaidhyaMegha Private Limited, India\\[2pt]\url{https://samyama.ai/}\\[8pt]July 2026}

\begin{document}
\maketitle

\begin{abstract}
Large language models pay a well-documented tax on non-English text: the same content costs several
times more tokens, and because attention is quadratic in sequence length, far more compute. We ask how
much of this tax is removable. Framing the token layer as source coding --
transformer compute is monotone in sequence length, whose per-atom floor is the Shannon rate
$H/\log_2 V$, an object already applied to tokenizers in prior work -- we assemble a token-cost ledger
that splits each language's cost, at fixed parallel content, into a removable coding redundancy, a
residual coding slack, an intrinsic-content term, and an orthogonal, irreducible grapheme-to-phoneme
term that governs the multimodal rather than the text cost. On FLORES-200 across eight languages, a
production tokenizer costs up to $8.9\times$ more tokens for Indic scripts than for English; a
script-matched code trained on $1{,}012$ sentences removes a median $64\%$ of that excess
(bootstrap 95\% CI $[0.638, 0.647]$), and a script-fair information floor shows the intrinsic content
differs by under $6\%$ -- the tax is representational, not informational. A constructed code removes
$98\%$ of a controlled source's redundancy, and the token tax implies up to $79\times$ attention cost.
We are explicit about scope and failure: this is compute-and-memory accounting, not a model-quality
claim; we neither measure nor claim the cross-lingual direction of the orthographic term; and our
matched code is a conservative small-data demonstration. We contribute the unifying ledger, the
removable-versus-intrinsic attribution, and an open one-command harness.
\end{abstract}

\section{Introduction}
It costs more to say the same thing to a language model in Telugu than in English. On identical,
professionally-translated content, a widely-used production tokenizer emits up to $8.9\times$ as many
tokens for an Indic script as for English (Section~\ref{sec:results}); an older English-centric
tokenizer reaches $16$--$20\times$ per word. Because self-attention is $\Theta(N^2)$ in sequence
length $N$ \citep{vaswani2017attention}, a $k\times$ token inflation is a $k^2\times$ attention-compute
inflation and a $k\times$ shrink of the effective context window. This multilingual ``token tax'' is
well documented as a phenomenon and a fairness problem \citep{ahia2023allcost, petrov2023unfairness,
lundin2025tokentax}.

The question this paper asks is narrower and, we think, more useful: \emph{how much of the tax is
removable by choosing a better code, and how much is intrinsic to the language?} We answer it by
assembling a \emph{token-cost ledger}. Holding semantic content fixed with a parallel corpus, we
decompose each language's normalized sequence length into (i) a \emph{removable} coding redundancy --
the penalty for using a code fit to the wrong (English-dominated) distribution; (ii) a residual coding
slack from a finite, finitely-trained vocabulary; (iii) an \emph{intrinsic}-content term measured by a
script-fair compressor; and, on an orthogonal axis, (iv) an irreducible grapheme-to-phoneme term that
does not affect text sequence length at all but governs the multimodal (speech) ledger. Terms (i)--(iii)
live on the text-compute axis; (iv) is the organic cost of a deep orthography.

None of the ledger's individual pieces is new. The fertility floor $F^\star=H/\log_2 V$ is Shannon
source coding \citep{shannon1948mathematical, cover2006elements}, applied to tokenizers as an
``efficiency'' or ``capacity-utilization'' metric by \citet{zouhar2023noiseless} and
\citet{erdogan2026infotok}, the latter of whom also give the additive floor-plus-redundancy split we
use. Orthographic depth as an information-theoretic quantity is established
\citep{torres2025orthography}. A compute-optimal vocabulary size is derived by
\citet{tao2024vocabulary}. Our contribution is (a) to \emph{unify} the removable text term and the
irreducible orthographic term into one cost ledger; (b) to \emph{attribute} the observed multilingual
tax empirically to removable versus intrinsic components on real parallel data, with a constructed code
that drives the removable term to zero; and (c) to keep the whole thing scoped strictly to
\emph{compute and memory}, which is what makes the accounting well-posed.

\paragraph{Scope fence (stated once, up front).} This is a cost accounting, not a quality claim. We do
\emph{not} assert that fewer tokens make a better model; \citet{schmidt2024morethan} and others show it
need not, and we agree. Everything here concerns FLOPs, KV-cache bytes, and context-window occupancy,
quantities that are monotone in sequence length regardless of downstream accuracy.

\paragraph{Contributions.}
\begin{itemize}[leftmargin=1.4em,itemsep=1pt]
\item \textbf{A unified token-cost ledger} (Section~\ref{sec:model}) that puts the removable coding
redundancy and the irreducible orthographic term on one accounting identity, with a single estimand --
the removable fraction $\rho$ -- for ``how synthetic is this language's tax?''
\item \textbf{An empirical attribution} (Section~\ref{sec:results}) on FLORES-200: a script-matched code
trained on $1{,}012$ sentences removes a median $\rho=0.64$ of the production-tokenizer tax (CI
$[0.638,0.647]$), and the script-fair information floor shows intrinsic content varies by $<6\%$ across
Indic languages -- the tax is representational.
\item \textbf{A constructed floor-approaching code} and the quadratic-cost consequence
(Section~\ref{sec:law}): a matched code removes $98\%$ of a controlled source's redundancy, sitting
$0.036$ bits above the entropy floor; the token tax implies up to $79\times$ attention cost.
\item \textbf{An open, one-command harness} and an honest limitations section
(Section~\ref{sec:limits}), including a pre-registered control that we report as \emph{not} established.
\end{itemize}

\section{The token-cost ledger}\label{sec:model}
\paragraph{Cost is monotone in sequence length.} A decoder-only transformer of width $d$ and $L$ layers
processing $N$ tokens pays $\Theta(L N^2 d)$ attention FLOPs, $\Theta(L N d^2)$ feed-forward FLOPs,
$\Theta(L N d)$ KV-cache memory, and an $O(Vd)$ vocabulary term \citep{vaswani2017attention}.

\begin{proposition}[cost monotonicity]\label{prop:mono}
For fixed $(d,L,V)$, transformer compute and KV memory for a fixed piece of content are non-decreasing
in $N$, and strictly increasing once $N\gtrsim d$ (the attention term dominates).
\end{proposition}
\noindent Every term is non-decreasing in $N$; the $N^2$ term's derivative $2LNd$ overtakes the linear
terms once $N>O(d)$. The consequence is the objective: \emph{minimize expected $N$ for fixed content}.
In the quadratic regime a sequence-length ratio $r$ is an attention-cost ratio $r^2$.

\paragraph{The floor (prior art).} Model content as atoms drawn from a source $p$ with entropy $H(p)$
bits/atom, encoded to tokens over a $V$-ary alphabet.
\begin{lemma}[fertility floor; \citealp{cover2006elements, zouhar2023noiseless, erdogan2026infotok}]
\label{lem:floor}
Any uniquely-decodable $V$-ary code has expected tokens per atom $F \ge H(p)/\log_2 V =: F^\star$, with
equality approached within one token by a matched Huffman code and exactly in the block limit. If the
code is optimal for a wrong distribution $q$, then $F = F^\star + R$ with $R = D_{\mathrm{KL}}(p\,\|\,q)/\log_2 V \ge 0$.
\end{lemma}
\noindent We state Lemma~\ref{lem:floor} for self-containedness and \emph{attribute} it; $\eta=H/\log_2 V$
is \citeauthor{erdogan2026infotok}'s capacity utilization, and the $F^\star+R$ split is their
Appendix-C decomposition. $R$ is the \emph{removable} part: refit the code to $p$ and $R\to0$.

\paragraph{The ledger (our synthesis).} Hold content fixed with a parallel corpus and normalize every
quantity to a reference language (English $=1$). Write $\mathrm{NSL}_E(L)$ for the total tokens of
language $L$ under encoder $E$ divided by English's under the same encoder. Then the excess of a
production encoder decomposes additively into individually measurable terms that are non-negative for
every language whose tax we attribute:\footnote{The identity telescopes exactly for all languages; the
residual-slack term $\mathrm{NSL}_{\mathrm{match}}-\mathrm{NSL}_{\mathrm{floor}}$ can nonetheless turn
marginally negative for a shallow-orthography Latin control whose intrinsic content already sits near
the English baseline. German is the one such case here ($\mathrm{NSL}_{\mathrm{match}}=1.13<
\mathrm{NSL}_{\mathrm{floor}}=1.15$): the small matched BPE and the LZMA content proxy are distinct
instruments, each normalized to English under itself, and agree to within noise this close to $1$. For
all five Indic languages---our focus---every term is strictly positive (Table~\ref{tab:ledger}).}
\begin{equation}\label{eq:ledger}
\underbrace{\mathrm{NSL}_{\mathrm{prod}}(L)-1}_{\text{observed tax}}
= \underbrace{[\mathrm{NSL}_{\mathrm{prod}}-\mathrm{NSL}_{\mathrm{match}}]}_{\text{removable }R\text{ (vocab mismatch)}}
+ \underbrace{[\mathrm{NSL}_{\mathrm{match}}-\mathrm{NSL}_{\mathrm{floor}}]}_{\text{residual coding slack}}
+ \underbrace{[\mathrm{NSL}_{\mathrm{floor}}-1]}_{\text{intrinsic content}},
\end{equation}
and, on an orthogonal axis, an irreducible term $G(L)=H(\text{phoneme}\mid\text{grapheme})$ that leaves
text $N$ untouched but governs the speech ledger. The unification of the removable text term with the
irreducible orthographic term is the contribution; neither half is ours (the removable term is
\citealp{erdogan2026infotok}; orthographic depth as algorithmic mutual compressibility is
\citealp{torres2025orthography}, whose Kolmogorov instrument we replace with a Shannon one in the
LLM-cost setting). The headline estimand is the \emph{removable fraction}
\begin{equation}\label{eq:rho}
\rho(L)=1-\frac{\mathrm{NSL}_{\mathrm{match}}(L)-1}{\mathrm{NSL}_{\mathrm{prod}}(L)-1},
\end{equation}
the share of the production tax a language-matched code removes: $\rho\!\to\!1$ means the tax is a code
artifact (synthetic), $\rho\!\to\!0$ means it is intrinsic (organic).

\paragraph{A one-line remark on vocabulary.} Substituting $N=A\,H/\log_2 V$ into the cost model gives a
$\mathrm{Cost}(V)$ with an interior minimizer $V^\star$ balancing shrinking sequence terms against a
growing vocab term -- but this compute-optimal vocabulary is already established, empirically, by
\citet{tao2024vocabulary} and \citet{limisiewicz2026compute}; we cite it and claim nothing here.

\section{Experimental setup}\label{sec:setup}
\textbf{Real parallel data.} FLORES-200 \citep{nllb2022flores} (CC-BY-SA), professionally translated and
sentence-aligned, so line $i$ of every language file is the same sentence. We report on eight
languages: a Latin control (English, Spanish, German), Devanagari (Hindi), Bengali, and three Dravidian
abugidas (Telugu, Tamil, Kannada). We measure on the $1{,}012$-sentence \texttt{devtest} split.
\textbf{Encoders.} UTF-8 bytes; Unicode extended grapheme clusters (\texttt{\textbackslash X}, the
akshara-preserving atom); three production byte-level BPE tokenizers (GPT-2, \texttt{cl100k\_base},
\texttt{o200k\_base}); and a per-language \emph{matched} BPE trained \emph{held out} on the FLORES
\texttt{dev} split and evaluated on \texttt{devtest}. \textbf{Information floor.} A script-fair content
estimate: LZMA over the grapheme-cluster-ID stream (not UTF-8 bytes), so a script is not charged for its
3-byte-per-codepoint UTF-8 assignment. \textbf{Constructed source.} A Zipf($s{=}1.1$) source over $256$
concepts with a known entropy. \textbf{Apparatus.} Single-thread Python; \texttt{tiktoken} and the
standalone \texttt{tokenizers} library (no GPU, no model weights); deterministic seeds; one command
regenerates every number and figure.

\paragraph{Pre-registration.} Hypotheses, decision rules, and four negative controls were frozen before
the confirmatory run. The calibration control (NC2) requires the stack to recover a \emph{known}
entropy: on a uniform-32 source it returns $\hat H=5.0000$ bits and a Huffman length in $[H,H{+}1)$;
it passed before any language number was read.

\section{Results: the removable tax}\label{sec:results}
\begin{figure}[!ht]\centering
\includegraphics[width=0.72\linewidth]{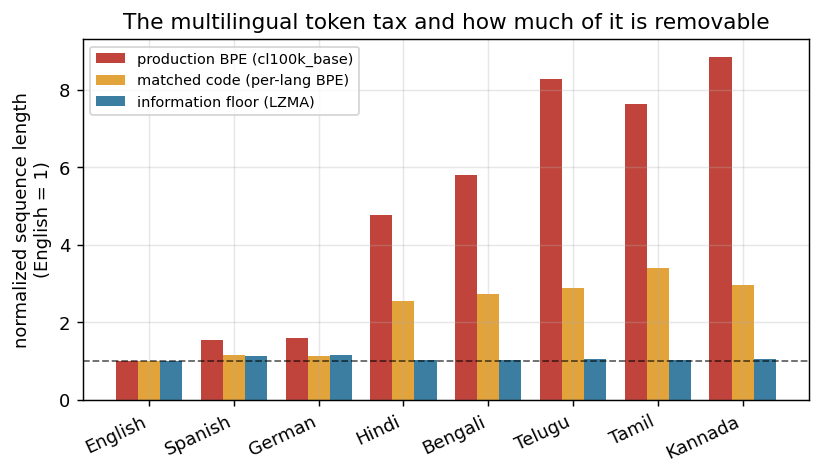}
\caption{The multilingual token tax and how much of it is removable. Production BPE (red) vs.\ a
script-matched code (orange) vs.\ the script-fair information floor (blue), normalized to English.}
\label{fig:tax}
\end{figure}

Table~\ref{tab:ledger} and Figure~\ref{fig:tax} report the ledger. Under \texttt{cl100k\_base} the
production tax reaches $8.29\times$ (Telugu), $8.86\times$ (Kannada), and $4.76\times$ (Hindi); under the
older GPT-2 tokenizer, fertility per word reaches $16.4$--$20.3\times$ for Dravidian scripts. A
\emph{script-matched} code trained on only $1{,}012$ sentences (learned vocabulary $2{,}000$--$3{,}000$)
brings these to $2.89$, $2.95$, and $2.54\times$; and the script-fair information floor sits at
$1.02$--$1.06\times$ -- the intrinsic content of the same sentences is within $6\%$ across all Indic
languages (this is our pre-registered content-invariance control, NC3: no language exceeded the $1.5\times$
threshold, so no part of the tax is re-attributed to intrinsic content). Bootstrapping sentences, the median removable fraction across the five Indic languages is
$\rho=0.642$ with a $95\%$ CI of $[0.638,0.647]$: \textbf{a matched code removes about two-thirds of the
production tax, and the information floor shows nearly all of the rest is coding slack rather than
content}. That production tokenizers themselves disagree by $4\times$ on the same Telugu content
($8.29\times$ under \texttt{cl100k} vs.\ $1.93\times$ under the more multilingual \texttt{o200k}) is
independent evidence that the tax is a property of the code, not the language.

\begin{table}[t]\centering\small
\begin{tabular}{lrrrrr}
\toprule
Language & bytes/char & $\mathrm{NSL}_{\mathrm{prod}}$ & $\mathrm{NSL}_{\mathrm{match}}$ & $\mathrm{NSL}_{\mathrm{floor}}$ & $\rho$ \\
\midrule
English & $1.00$ & $1.00$ & $1.00$ & $1.00$ & --- \\
Spanish & $1.02$ & $1.54$ & $1.15$ & $1.12$ & $0.72$ \\
German  & $1.02$ & $1.59$ & $1.13$ & $1.15$ & $0.78$ \\
Hindi   & $3.92$ & $4.76$ & $2.54$ & $1.02$ & $0.59$ \\
Bengali & $4.22$ & $5.80$ & $2.72$ & $1.02$ & $0.64$ \\
Telugu  & $4.62$ & $8.29$ & $2.89$ & $1.04$ & $0.74$ \\
Tamil   & $4.23$ & $7.64$ & $3.39$ & $1.04$ & $0.64$ \\
Kannada & $4.15$ & $8.86$ & $2.95$ & $1.06$ & $0.75$ \\
\midrule
\textbf{Indic median} & & & & & \textbf{0.64}\ \tiny{CI[0.638,0.647]} \\
\bottomrule
\end{tabular}
\caption{The token-cost ledger on FLORES-200 \texttt{devtest} ($1{,}012$ parallel sentences), production
tokenizer \texttt{cl100k\_base}. $\mathrm{NSL}$ = normalized sequence length (English $=1$).
$\mathrm{NSL}_{\mathrm{floor}}$ is the script-fair LZMA content ratio. $\rho$ is the removable fraction
(Eq.~\ref{eq:rho}); the matched code is trained held-out on $\sim$1k sentences (a conservative
demonstration).}
\label{tab:ledger}
\end{table}

\paragraph{Bytes-per-char predicts the tax (H2, a replication).} Across the eight languages, UTF-8
bytes-per-character correlates with the production tax at Spearman $\rho_s=0.83$ ($p=0.01$), replicating
\citet{ahia2023allcost} and \citet{petrov2023unfairness}; our addition is the decomposition, not the
correlation.

\FloatBarrier
\section{Decomposition, a constructed code, and quadratic cost}\label{sec:law}
\begin{figure}[!ht]\centering
\includegraphics[width=0.72\linewidth]{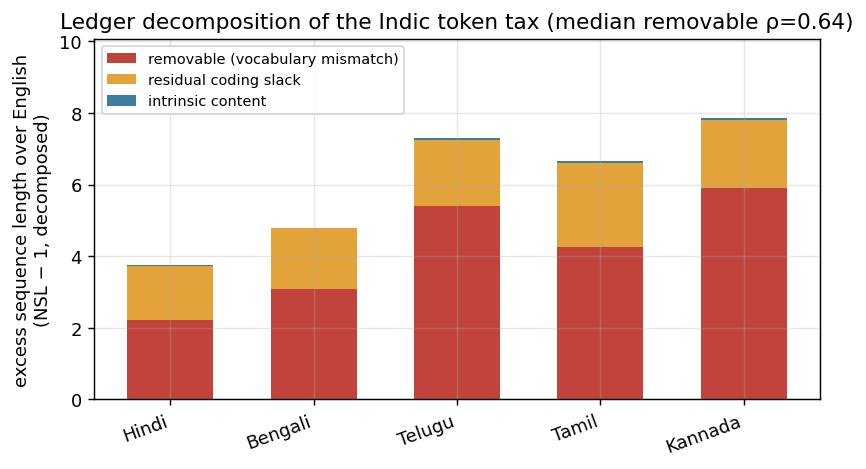}
\caption{Ledger decomposition of the Indic token tax (Eq.~\ref{eq:ledger}): removable coding redundancy
(red) dominates; residual coding slack (orange) is removable in principle; intrinsic content (blue) is a
sliver.}
\label{fig:decomp}
\end{figure}

Figure~\ref{fig:decomp} decomposes the Indic excess of Eq.~\ref{eq:ledger}. The removable term
(vocabulary mismatch) dominates; the intrinsic-content sliver is $\le 0.06$ in every case. The residual
coding slack -- the gap between our small matched code and the floor -- is itself removable in principle
(a better-trained code closes it), so it is a lower bound on removability, not a second intrinsic term;
that a $200$k-vocabulary production tokenizer (\texttt{o200k}) already reaches $1.93\times$ on Telugu
confirms the slack is training, not content.

\begin{figure}[!ht]\centering
\includegraphics[width=0.92\linewidth]{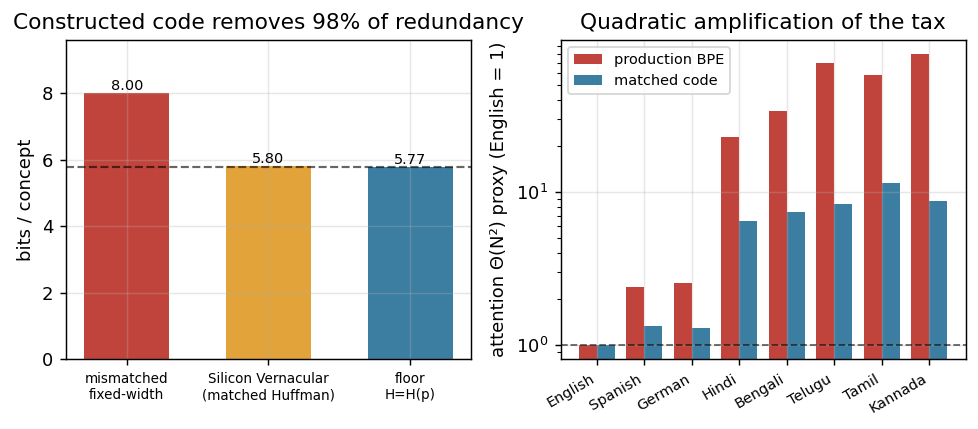}
\caption{Left: a constructed matched code removes $98\%$ of a controlled source's redundancy, landing
$0.036$ bits above the entropy floor. Right: the $\Theta(N^2)$ attention-cost proxy (English $=1$),
production vs.\ matched -- the token tax is quadratically amplified in compute.}
\label{fig:constructed}
\end{figure}

\paragraph{A constructed code hits the floor.} On the controlled Zipf source (entropy $H=5.77$
bits/concept), a mismatched fixed-width code pays $8.0$ bits/concept; the matched Huffman code
\citep{huffman1952method} (``Silicon Vernacular'') pays $5.80$ -- $0.036$ bits above the floor, within Lemma~\ref{lem:floor}'s
one-bit guarantee -- removing $98\%$ of the redundancy (Figure~\ref{fig:constructed}, left). Our pre-registered no-sub-floor control (NC1) requires that no
code we report encodes below $H$, and the Kraft sum of every code we build satisfies $\sum 2^{-\ell_i}\le 1$. We
therefore deliberately do \emph{not} report an empirical block code beating the floor: at block sizes $>1$ the
$k$-gram alphabet is undersampled and an in-sample Huffman would appear to beat $H$ from finite-sample
bias -- a dishonest number. The residual sub-bit gap is closed by block/arithmetic coding as a theorem,
not a measurement. We position this construction as the discrete-code limit of byte-entropy patching
\citep{pagnoni2024blt}, not as a proposed human language.

\paragraph{Quadratic amplification.} Because attention is $\Theta(N^2)$, the token tax is amplified in
compute: the same content costs up to $79\times$ (Kannada) and $69\times$ (Telugu) the attention work of
English under \texttt{cl100k}, collapsing to $8$--$11\times$ under the matched code
(Figure~\ref{fig:constructed}, right).

\FloatBarrier
\section{Related work}\label{sec:related}
\textbf{Information theory of tokenizers.} \citet{zouhar2023noiseless} frame the tokenizer as a channel
and define an entropy-to-max-entropy efficiency (the parent of $H/\log_2 V$); \citet{erdogan2026infotok}
define capacity utilization $\eta=H_1/\log_2 K$ and the additive floor-plus-redundancy decomposition we
build on; \citet{rajaraman2024theory} and \citet{gastaldi2025foundations} give complementary theories of
tokenization. We contribute neither the floor nor the split -- we contribute their \emph{use} as a
multilingual accounting. \textbf{The multilingual tax.} \citet{ahia2023allcost},
\citet{petrov2023unfairness}, and \citet{lundin2025tokentax} document the cross-language cost disparity;
Indic-specific tokenizers \citep{indicsupertok2025} reduce fertility. We explain the disparity as a
removable KL redundancy above a script-fair floor -- and note that \citeauthor{petrov2023unfairness}'s
residual byte-level disparity is exactly the intrinsic + orthographic remainder our ledger predicts.
\textbf{Byte-level models.} MEGABYTE \citep{yu2023megabyte} and the Byte Latent Transformer
\citep{pagnoni2024blt} remove the discrete code and let entropy set patch boundaries; our ledger
\emph{explains} what they can remove (the redundancy $R$) and cannot (intrinsic content, orthographic
$G$). \textbf{Compression is not quality.} \citet{schmidt2024morethan} and \citet{bostrom2020bpe} show
fewer tokens need not improve models; our scope fence makes this orthogonal -- we account for cost, not
quality. \textbf{Orthographic depth.} \citet{torres2025orthography} formalize transparency as Kolmogorov
mutual compressibility; we borrow the quantity (as Shannon $H(\text{phoneme}\mid\text{grapheme})$) for
the irreducible axis. \textbf{Optimal vocabulary.} \citet{tao2024vocabulary} and
\citet{limisiewicz2026compute} derive the compute-optimal vocabulary/granularity we merely cite.

\section{Limitations and honest negatives}\label{sec:limits}
(1) \textbf{We did not establish the orthographic direction.} We pre-registered a control (NC4) that
English's grapheme-to-phoneme ambiguity exceeds shallow Indic scripts'. We measure only the English side
(CMUdict homograph entropy $0.070$ bits/type); we have no Indic pronunciation lexicon, so the
cross-lingual direction is literature-consistent but \emph{not} established by our instrument. Per the
pre-registration we report this as a negative and future work, not a result. (2) \textbf{The matched code
is a small-data demonstration}, trained on $\sim$1k sentences (vocabulary $2$--$3$k); it understates
removability, making $\rho=0.64$ a lower bound. (3) \textbf{The information floor is an LZMA estimate},
an upper bound on intrinsic content; a tighter estimator would shrink the intrinsic term further, again
in the direction of ``more removable.'' (4) \textbf{This is not a quality claim.} We measure compute and
memory; we make no statement about accuracy or loss, and explicitly do not contradict
\citet{schmidt2024morethan}. (5) \textbf{Scope}: eight languages, one parallel benchmark, text only;
speech/multimodal cost is argued, not measured. (6) We prove no new theorem: the floor and its
redundancy split are cited, not claimed.

\section{Conclusion}
The multilingual tokenization tax is, in the compute ledger, mostly \emph{removable}: on real parallel
text a script-matched code trained on a thousand sentences erases about two-thirds of it, and a
script-fair floor shows the intrinsic content of the same sentences differs by under six percent. What
remains -- the organic grapheme-to-phoneme cost of a deep orthography -- is real but lives on a different
(multimodal) axis, and flips the ranking. We contribute the unifying ledger, the removable-versus-intrinsic
attribution, a constructed code that reaches the entropy floor, and an open harness; we claim neither the
floor, nor the optimal vocabulary, nor a model-quality benefit. The larger program these results open --
\emph{design principles of a near-optimal language for foundational models}, spanning grammar and
attention routing, acoustic isomorphism, and human learnability -- we name as future work and a thesis,
not a claim of this paper.

\paragraph{Reproducibility.} Code, data pointers, pre-registration, and one-command reproduction:
\url{https://github.com/samyama-ai/token-cost-ledger}.

\small
\bibliographystyle{plainnat}
\bibliography{paper20_token_cost_ledger}
\end{document}